\def\year{2019}\relax
\documentclass[letterpaper]{article} 
\usepackage[hyperref]{aaai19}  
\usepackage{times}  
\usepackage{latexsym}
\usepackage{helvet}  
\usepackage{courier}  
\usepackage{url}  
\usepackage{amsmath}
\usepackage{graphicx,caption,subcaption}  
\usepackage{amssymb}
\usepackage{adjustbox}
\usepackage{bm}

\begin{document}

\title{Transferable Natural Language Interface to Structured Queries \\ aided by Adversarial Generation}
\author{Hongyu Xiong \\
  Stanford University, CA 94305 \\
  {\tt hxiong2@stanford.edu} \\
  \And
  Ruixiao Sun \\
  Stanford University, CA 94305 \\
  {\tt ruixiaos@stanford.edu} \\}

\maketitle

\begin{abstract}
A natural language interface (NLI) to structured query is intriguing due to its wide industrial applications and high economical values.
In this work, we tackle the problem of domain adaptation for NLI with limited data on target domain.
Two important approaches are considered: (a) effective general-knowledge-learning on source domain semantic parsing, and (b) data augmentation on target domain. 
We present a Structured Query Inference Network (SQIN) to enhance learning for domain adaptation, 
by separating schema information from NL and decoding SQL in a more structural-aware manner; 
we also propose a GAN-based augmentation technique (AugmentGAN) to mitigate the issue of lacking target domain data.
We report solid results on \textsc{GeoQuery}, \textsc{Overnight}, and \textsc{WikiSQL} to demonstrate state-of-the-art performances for both in-domain and domain-transfer tasks.
\end{abstract}


\section{Introduction}
\label{sec:introduction}

Great efforts have been invested in deep-learning-based semantic parsing to convert natural language (NL) texts to structured representation or logical forms \cite{Wang2015BuildingAS,PasupatL15,Jia2016DataRF}. 
In particular, a special case of semantic parsing 
- natural language interface (NLI) to structured query like SQL \cite{androu1995natural,popescu2003towards,li2005nalix,li2014nalir} 
- has incited significant interests; 
the motivation is two-fold: 
1) majority data in the world is stored in relational tables (databases), 
and a NLI to database engine has the potential to support a great many dialogue-based applications; 
2) it is extremely difficult for machine to understand the meanings of arbitrary NL texts, especially involving multiple different domains,
but the complexity is more likely to be reduced by converting texts to formal languages.

Some previous works have been using seq2seq \cite{sutskever2014sequence} models to generate structured queries for a certain database, given NL queries \cite{Dong2016,Jia2016DataRF}. 
Provided with abundant data and through an end-to-end training process, seq2seq model achieves decent performance on a single relational table; 
however, it is not straightforward to apply a trained model to a new table. 
For example, suppose we have two queries $Q_1$ and $Q_2$ against a geography table and an employee table, respectively:
\begin{equation*}
    \begin{array}{llll}
    \small Q_1: \textrm{what's the \textbf{size} of \textit{south america}?}
    \\
    \small Q_2: \textrm{what's the \textbf{age} of \textit{john smith}?}
    \end{array}    
\end{equation*}
a model trained for the geography table is able to parse $Q_1$, but when if comes to the employee table, the model would fail to directly parse $Q_2$. 

This is because seq2seq model with end-to-end training mixes up three types of knowledge: 
(a) the ontology of NL (grammar), (b) domain-specific language usage, and (c) the schema information of the relational table (column and value). 
Back to the example, the end-to-end model has only trained on ``age'' but not ``size'', ``john smith'' but not ``south america'', 
so even though they are both schema information, it is difficult to use a model trained for one table (source domain) to answer queries against another (target domain). 

Therefore, for a reliable domain-adaptation solution, two important approaches should be considered: 
(1) improve the learning of the NL ontology knowledge on source domain, which could then extend to target domain; 
and (2) manage to augment more data on target domain, so general NL knowledge and domain-specific usage are better learned. 
In this paper, we address the two problems accordingly:  
\begin{enumerate}
\item We design a Structured-Query Inference Network (SQIN) for better cross-domain semantic-parsing, by separating schema-related information from NL query and decoding SQL in a more structural-aware way;
\item We design a generative adversarial network AugmentGAN to augment the limited number of training data on target domain.
\end{enumerate}

\section{General Approaches}
\label{sec:assum-appro}

To make domain adaptation more effective and with less required resources, we are going to: 
(1) explicitly separate the relational-table-related information in the NL query and structurally generate SQL, 
and (2) given limited number of target domain data, try to effectively augment to a larger size. 
We will first introduce the scope of our method and explain its validity.

\subsection{Assumptions}
\label{sec:expressive-power}
1. \underline{Single relational table (self-join operations supported} \underline{as subquery):} This assumption implies that the SQL we support is a subset of standard SQL. 
A recent detailed analysis \cite{JohnsonNS18} reveals that $\sim 55\%$ of 8.1 million industrial SQL queries are against single relational tables with self-join; 
in real life people are more likely to inquire simple and structured data such as weather or stock prices, 
so it is fair to say the percentage of assumed types of queries is even higher in most practical NL-based applications. 
Our method is also capable of extending to more complex cases.

2. \underline{Column names (and corresponding types) of a table} \underline{are provided only:} In most circumstances for privacy concerns, values stored in the table should not be accessible by NLI providers, unlike a new work \textit{STAMP} \cite{Sun2018SemanticPW}, where value/cell could be accessible. 
During domain transfer, in addition to the schema, a limited number of (NL, SQL) pairs on target domain are given for training. 

3. \underline{Column/value information is explicitly mentioned:} This assumption ensures that we could identify and match the columns and values against an NL query. 
We do not require the columns to match exactly their appearances in the table;
different forms (like plurals or past tense), synonyms, and common usages are allowed. 
For examples,
\begin{equation*}
    \begin{array}{llll}
    \small Q_1': \textrm{how \textbf{large} is \textit{south america}?}
    \\
    \small Q_2': \textrm{how \textbf{old} is \textit{john smith}?}
    \end{array}    
\end{equation*}
are also dealt with by our method.
In this paper, NL queries which are too implicit will not be our focus.

\subsection{Domain-adaptive Semantic Parsing}
We present a Structured-Query Inference Network (SQIN), by dividing the semantic parsing task into two stages: 

(1) Tag the column names and value information against the NL input. 
Some existing works proposed to detect schema information and copy them directly to the output using an attention-copying mechanism \cite{Jia2016DataRF,vinyals2015pointer}; 
however, intensive learning is still needed when moving to another domain.
Here, we use a convolutional tagging network (CTN) to determine, for each token in the NL query, whether it is a column, a value of column, or \texttt{nan}. For example, suppose we have schema $S_1 =$ [`country',`size',`population'] for the geography table and $S_2 =$ [`name',`salary',`age'] for the employee table, then $Q_1$ and $Q_2$ will be tagged in the following forms:
\begin{equation*}
    \begin{array}{llll}
    \small T_1: \textrm{what's the \textbf{size}:\texttt{c2} of \textit{south america}:\texttt{v1}? }
    \\
    \small T_2: \textrm{what's the \textbf{age}:\texttt{c3} of \textit{john smith}:\texttt{v1}? } 
    \end{array}
\end{equation*}

(2) Convert the tagged NL query to SQL query, e.g. $T_1$ and $T_2$ will be converted to SQL formats:
\begin{equation*}
    \begin{array}{llll}
    \small L_1: \textrm{select \texttt{c2} where \texttt{c1} = \texttt{v1}}
    \\
    \small L_2: \textrm{select \texttt{c3} where \texttt{c1} = \texttt{v1}}
    \end{array}
\end{equation*}
where a \texttt{FROM} statement is omitted. In the end, the column tags are substituted by the column names in the schema, and the value tags substituted by the corresponding substrings from the input.

For more complex SQLs, to decode in a more structural-aware manner, both the (a) hierarchical and (b) compositional properties of SQL queries should be addressed accordingly. 
\textit{SQLNet} \cite{xu2017sqlnet} uses a Seq2Set model with a sketch to deal with the compositional nature of SQL, 
but it only works for simple types of SQL and requires a significant amount of human efforts to define and retrain a new sketch, 
making it hard to adapt to different types of SQL. 
\textit{Seq2Tree} \cite{Dong2016} tackles the hierarchical structure of SQL by using a hierarchical tree decoder, 
but it still requires the model to memorize different possible compositions of keywords.
\textit{ASN} \cite{rabinovich2017abstract} incorporated both tree-like structure and recursive decoding, but the multi-module design could be significantly simplified.
Therefore, we use a simple structured sequence-based parts-of-SQL (\textit{seq2PSQL}) generation to capture both natures of SQL with the help of the tagged information from the NL query.

More details of the model design will be introduced in the later sections.

\subsection{Augment Data on Target Domain}
\label{sec:discuss_augment}
Jia and Liang (2016) \cite{Jia2016DataRF} previously developed a recombination method for data augmentation; 
for example, their \textsc{AbsEntity} method replaces a value in the query with different values for the same column; 
their \textsc{AbsWholePhrases} method replaces a value with its column under certain conditions. 
However, it would appear that the model might be heavily biased by the small set of seed queries used to generate the query variations, 
which may cause the augmented set of NL queries simpler than the full scope of NL expression.

We propose an augmenting algorithm that goes beyond. Given two different NL queries, it is very difficult to hybridize parts of them and generate a  new fluent NL text; 
however, it is simple to recombine two SQL queries, as SQL is based on strict grammar rules. Therefore, we propose to train another sequence-based model to generate a NL query given certain SQL, 
and the augmentation process is to generate the corresponding NL queries for recombined SQL queries. 
We adopt a generative adversarial network (GAN), by using a discriminator to classify whether the generated NL query resembles the human usage, and the result is used as reward to the generator.

More details of the model design will be introduced in the later sections.

\section{Related Works}
\label{sec:related-work}

Seq2seq-based \cite{sutskever2014sequence} models enable semantic parser training in an end-to-end manner without manual feature engineering. 
Besides common seq2seq framework \cite{Jia2016DataRF,xiao2016sequence,zhong2017seq2sql}, there are other sequence-based models with structural-aware decoders like \textit{Seq2Tree} \cite{Dong2016}, \textit{SQLNet} \cite{xu2017sqlnet}, \textit{EG} \cite{wang2018robust} and \textit{Abstract Syntax Networks} \cite{rabinovich2017abstract}.
Due to the black-box nature of seq2seq, both Cheng et al. \cite{cheng2017learning} and \textit{Coarse2fine} \cite{Dong2018coarse} proposed two-stage semantic parsers with the 1st stage mapping utterances to intermediate states, and 2nd stage converting intermediate states to logical forms. 
\textit{STAMP} \cite{Sun2018SemanticPW} realizes the importance of “linking” between the question and the table columns, and adopts a a switching gate in decoder and include value/cell information in SQL generation.
A most recent work \textit{MQAN} \cite{mccann2018natural} designs a multi-pointer-generator decoder for the generation.
As another line of works in deep-learning-based semantic parsing for relational tables, \textit{Neural Enquirer} \cite{YinLLK15} proposes a fully distributed end-to-end model where all components (query, table, answer) are stored and differentiable, 
and \textit{Neural Programmer} \cite{Neelakantan2016} defines a set of symbolic operators; 
these approaches lack of explicit interpretation and adaptability to different tables, and the input will be executed to generate an answer, instead of a structured query. 
Other progresses like \textit{Neural Symbolic Machine} \cite{liang2017neural} adopts memory for seq2seq model, but this will not be our focus in this work.
Two cross-domain seq2seq approaches \cite{su2017cross,herzig2017neural} are relevant, but both require a large amount of target-domain data to achieve a good domain adaptation. One recent work \cite{xiong2018transfer}
has made a good attempt to separate the schema information from the natural language query through annotation. As a future direction, \textit{DialSQL} \cite{Gur2018DialSQL} incorporates user feedbacks to enhance generation.

The idea of GANs \cite{RadfordMC15,chen2016infogan,salimans2016improved} has recently enjoyed success in NLP fields \cite{lamb2016professor,yu2017seqgan}. 
For example, a success application of GAN is used in \textit{Neural Dialogue Generation} \cite{li2017adversarial}, where the generator is a RL-based seq2seq model, and the outputs from the discriminator are then used as rewards for the generator, pushing the system to generate dialogues that mostly resemble human usage.


\section{Structural Query Inference Network (SQIN)}
\label{sec:separation}

\begin{figure*}[ht]
\begin{center}
\includegraphics[width=1.0\textwidth]{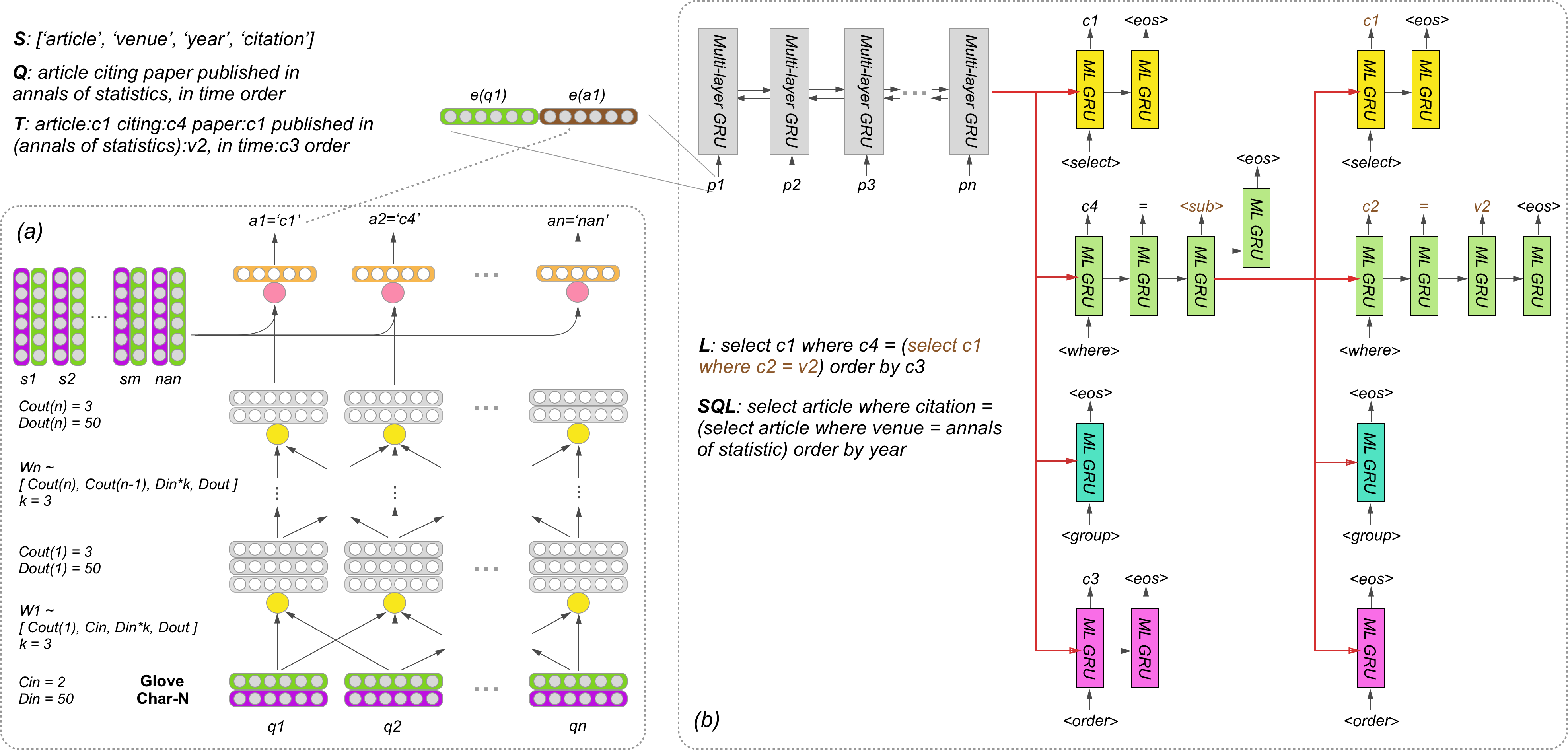}
\small
\caption{The working flow of SQIN. 
(a) CTN: For each token in $Q$, we use its Glove (green) and char-n-gram (purple) embeddings as input; at each layer the conv op is followed by a $tanh$ (yellow dot); the output of last layer multiplies with the $m+1$ embeddings from the schema $S$ through a bilinear matrix, followed by a softmax (red dot). The output $\{a_n\}$ are the column/value tags. 
(b) Seq2PSQL. The embeddings of each token and its tag are concatenated as the input $\{p_n\}$ of bi-encoder; with different starting tokens, the encoded hidden state generates different parts of SQL query $L$, and \texttt{<sub>} hierarchically launches a subquery generation. }
\label{Seq2Clauses}
\end{center}
\end{figure*}

In this section, we tackle the problem of domain-adaptive semantic parsing. 
Given the NL query $q = q_1, q_2, ..., q_n$, and the schema $S = [s_1, s_2, ..., s_m]$ of a relational table the query is against, our goal is to convert $q$ to corresponding SQL $l = l_1, l_2, ..., l_t$.

\subsection{Convolutional Tagging Network (CTN)}
\label{sec:DRN}
As discussed in \textbf{General Approaches}, first we want to identify and tag the columns and values information against the NL input with a sequence of tags $a = a_1, a_2, ..., a_n$, 
i.e. for each token $q_i$, we predict it with a tag $a_i$ denoting it as a column \texttt{cj}, value \texttt{vk}, or \texttt{nan}: \begin{equation}
f(q_1, q_2, ..., q_n) = (a_1, a_2, ..., a_n)
\end{equation}
where $a_i \in \{\texttt{cj}\}_m+\{\texttt{vk}\}_m+\{\texttt{nan}\}$ and both $j,k \in \{1,2,...,m\}$.

One challenge for the tagging is that a column name or a value could possibly consist of multiple tokens, 
so the model should capture the feature of neighboring tokens as well; therefore, we use a convolutional model. 
Another challenge is to choose suitable embeddings to represent the tokens: 
to capture a token with both semantic and character-level accuracies, we use both GloVe embedding \cite{pennington2014glove} and char-n-gram embedding \cite{kim2016character}, and regard them as two separate `channels' of this token.
For the embedding of a column name (multiple tokens considered), we take a bi-directional GRU to encode the two-channel word vector of each token \cite{zhong2017seq2sql}.

We use the multi-layer convolutional operations to process the NL query and assign tag for each token. 
For each conv layer, the input is the concatenation of $k$ consecutive $D_{in}$-dimensional embeddings with $C_{in}$ channels, and the output is a $D_{out}$-dimensional embeddings with $C_{out}$ channels, followed by a $tanh$ function; 
the convolution filter is with size of $(C_{out}, C_{in}, kD_{in}, D_{out})$.

For the last layer, each output is multiplied with the $m+1$ embeddings of the schema (plus \texttt{nan}) through a bilinear matrix with size $(D_{out}, D_{out})$; 
the $(1, m+1)$ result goes through a softmax function to return a $(1, m+1)$ probability vector; 
the index with the highest probability is related to one of the columns or \texttt{nan}. 
We call this model convolutional tagging network (CTN). 
Practically, we first use one CTN to tag column name against the NL input, and then add extra layers for values tagging.

\subsection{Sequence-based Parts of SQL (seq2PSQL) generation}
\label{sec:seq2clauses}
To encode the tagged NL query in the previous section, we use a sequence of embeddings $P = p_1, p_2, ..., p_n$, 
where $p_i = [e(q_i);e(a_i)]$ as a concatenation of original token $q_i$'s GloVe vector and its tag $a_i$'s embeddings. 
The tag embedding $e(a_i)$ itself is concatenated by three parts: 
(1) the embedding of tag type (\texttt{column} or \texttt{value}),
(2) the embedding of index $i$, which indicates the tag is either the $i$th column, or a value of the $i$th column, 
(3) the embedding of the value type (integer, string, etc).
Tags that share similar attributes (like same tag type or same id) also share part of embeddings. 

$P$ is taken as the input to a bidirectional multi-layer GRU encoder and encoded to a hidden representation. 
Since the encoder and decoder share the same vocabulary for the tags,
we use the same embedding for tags on both sides, 
and synchronize the updates of this embedding during back-propagation.

The decoder adopts a uni-directional multi-layer GRU and generates SQL queries in a top-down manner:

(1) to address the compositional nature of SQL, we use different starting tokens (like \texttt{<select>}, \texttt{<where>}...) 
to generate different clauses of SQL;
for each clause at each step, the output could be a column tag (\texttt{c1}), a value tag (\texttt{v2}), or a SQL functional word (like logical or aggregation operators);
generation terminates with an ending token \texttt{<eos>}; 
the decoders for different clauses share the same set of parameters, and by doing so all possible SQL clauses are adapted in one universal setting.

(2) to address the hierarchical nature, we define a nonterminal \texttt{<sub>} token which indicates the onset of a subquery. 
If \texttt{<sub>} is predicted, a new set of clauses start to decode by conditioning on the nonterminal's hidden vector. 
This process terminates when no more nonterminals are emitted.

In Fig.~\ref{Seq2Clauses}, we use an
example from \textsc{Overnight} \cite{Wang2015BuildingAS} to demonstrate the adaptive semantic parsing: a NL input is converted to a self-join SQL (supported as subquery).

\section{Data Augmentation based on GAN}
\label{sec:augmentGAN}
To augment the seed data, it is much easier to recombine SQLs and generate NL queries accordingly.
The problem can be framed as follows: given a SQL query $l$, the model needs to generate a NL query $q = q_1, q_2, ..., q_T$. 
We view the query generation as a sequence of actions taken according to a policy defined by a seq2seq model. 

\subsection{AugmentGAN}
In this section, we describe the proposed AugmentGAN model in detail. 

The adversarial paradigm is composed of a generator $G$ and a discriminator $D$. 
The key idea is to encourage $G$ to generate NL query that are indistinguishable from human,
using $D$ to provide reward for generation at each step. 
In detail, $G$ takes an attention-based seq2seq model \cite{bahdanau2014neural} that generates a NL query $q$ step by step given SQL $l$, 
and at each step, a partial query is generated and evaluated by $D$; 
$G$ is updated through reinforcement learning. 
$D$ is a binary classifier that takes a pair of SQL and NL queries ($l$, $q$) as input, 
and encodes into vector representations with size $H$ using two bi-directional GRU encoders, respectively; 
then the two hidden vectors are combined through a bilinear matrix with size $(H, H, M)$ to give a $(1, M)$ vector, 
which is fed to a 2-class feed-forward network, returning the score of being machine-generated or human-generated.

\begin{figure}[ht]
\begin{center}
\includegraphics[width=0.37\textwidth]{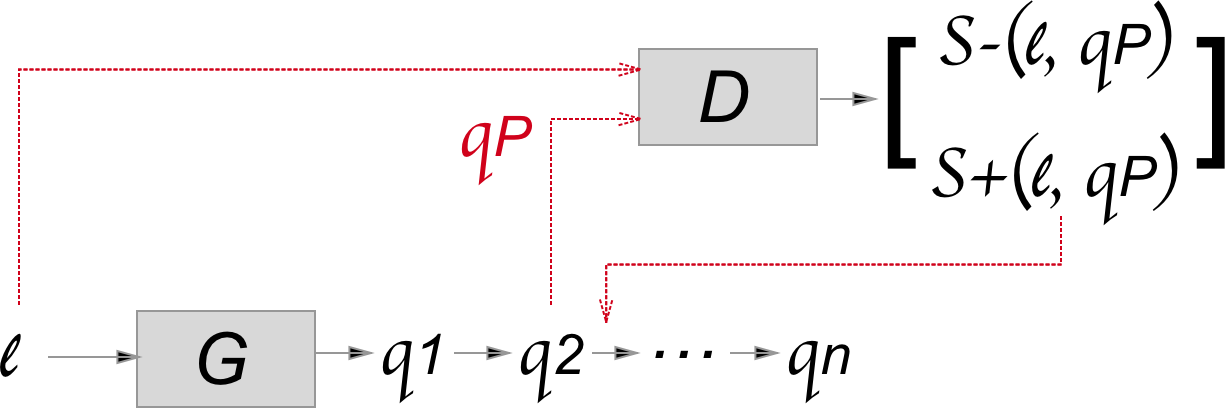}
\caption{The paradigm of AugmentGAN. Partially decoded sequence is passed into $D$ as reward for $G$ in the intermediate steps.
}
\label{fig:augment_gan}
\end{center}
\end{figure}

To calculate the score for the partial query $q_P$ at each step, we propose to use Monte Carlo Search \cite{li2017adversarial,yu2017seqgan}: 
the model keeps sampling tokens from the distribution until the decoding finishes, and repeats $N$ (set to $10$) times; 
we use the mean score of $N$ times sampling being human-generated ($S_{+}(\{l, q_P\})$) as the reward to update the policy of $G$ for the next step (Fig.~\ref{fig:augment_gan}).
The training objective is to maximize the expected reward of generated sequences based on policy gradient method \cite{williams1992simple}:
\begin{align*}
\nabla J(\theta) &\approx \Sigma_t [S_{+}(\{l, q_t\}) - b(l, q_t)] \\
& \nabla \textrm{log}p(q_t | x,q_{1:t-1})
\end{align*}
where $p(q_t | x,q_{1:t-1})$ is the policy of $G$, and $b(l, q_t)$ is the baseline function used to reduce the variance.

During the training, we also feed the human-generated query to the generator with a positive reward for model updates, which serves as a teacher intervene to give the generator more direct access to the gold-standard targets \cite{lamb2016professor,li2017adversarial}.


\section{Experiments \& Analyses}
We present experiment results on both in-domain and domain-transfer tasks, and also analyze our models and compare with previous works. 

\subsection{Datasets and Implementation}
\label{sec:dataset}
We train and evaluate our models on \textsc{GeoQuery} \cite{zettlemoyer2005learning}, \textsc{WikiSQL} \cite{zhong2017seq2sql}, and \textsc{Overnight} (sub-domain \textsc{Blocks} excluded for not being a relational table) \cite{Wang2015BuildingAS}. 
For data in \textsc{GeoQuery} and \textsc{Overnight}, we manually convert each original logical form to SQL query. 

For \textsc{GeoQuery} and \textsc{Overnight}, we use the standard train-test splits as released, and randomly divide the train sets to $5$ splits for model cross validation ($4:1$ for each train-valid cycle); the accuracies are calculated as the percentage of correct SQLs. For \textsc{WikiSQL}, we use the standard train-dev-test splits, and the accuracies are the percentage of correct logical forms (SQL queries).

We implement SQIN and AugmentGAN using Tensorflow, and train models using NVIDIA GPU card GTX-1080-Ti. For training, each iteration takes data with a batch size of 128, and the evaluation on development set happens for every 50 iterations. For \textsc{Overnight} dataset, it usually requires $5000 \sim 8000$ iterations for the models to achieve the performance shown in this work.

\subsection{In-domain Semantic Parsing}
For both CTN and seq2PSQL, we use pre-trained GloVe vector \cite{pennington2014glove} with dimension $D = 200$; 
for out-of-vocabulary (OOV) tokens not covered by GloVe, we randomly generate a vector using Gaussian distribution (with inferred element-wise mean and variance). 
The char-n-gram embeddings we use in CTN are pre-generated with $D = 200$ \cite{kim2016character}. 
The tag embedding is concatenated by three parts as discussed in previous section: (a) tag type, (b) id, and (c) value type, 
which are with dimension $100, 50, 50$, respectively, and randomly initialized using uniform scaling initializer $[ -\sqrt{3/D},  \sqrt{3/D}]$.

\begin{table}[ht]
\small
\begin{center}
\begin{adjustbox}{center}
\begin{tabular}{c|c|c|c}
\hline
Seq-based Models & \textsc{Geo.} & \textsc{Overn.} & \textsc{WkSQL} \\ \hline
seq2seq w/o RCB (Jia, 2016) & $.850$ & $.783$ & $.601$ \\
Seq2Tree (Dong, 2016) & $.871$ & $.808$ & $.624$ \\
Seq2SQL (Zhong 2017)  & $.854$ & $.786$ & $.604$  \\
ASN (Rabinovich, 2017) & $.871$ & \bm{$.812$} & $.629$ \\
SQLNet (Xu, 2017)  & - & - & $.680$ \\
STAMP w/o cell (Sun, 2018)  & - & - & $.674$ \\
Coarse2Fine (Dong, 2018)  & \bm{$.882$} & - & $.730$ \\
Execution-Guided (Wang, 2018)  & - & - & \bm{$.754$} \\
MQAN unordered (McCann, 2018)  & - & - & \bm{$.754$} \\
\hline
seq2PSQL  & $.875$   & $.815$ & $.668$ \\
CTN + seq2seq & $.908$ & $.819$ & $.724$ \\
SQIN & \bm{$.928$} & \bm{$.827$} & \bm{$.751$} \\
\hline
\end{tabular}
\end{adjustbox}
\end{center}
\caption{Test accuracies of SQLs on different datasets. \textsc{Overnight} subdomain \textsc{Blocks} is excluded. The seq2seq model \cite{Jia2016DataRF} is without augmentation, and the STAMP model \cite{Sun2018SemanticPW} is without cell information.}
\label{table:ablation}
\end{table}

\begin{table*}[ht]
\small
\begin{center}
\begin{adjustbox}{center}
\begin{tabular}{ l | l }
 \hline
 & \textbf{select article having count (venue) $\leq$ 2}  \\
 \hline
Ground Truth & article that has maximum two venues  \\
AugmentGAN & article with at most 2 venues  \\
Recomb. \cite{Jia2016DataRF} & article whose venues are at most two  \\
 \hline
 & \textbf{select author where article = (select article where (publish date) = 2004)}  \\
 \hline
Ground Truth & authors of articles published in 2004 \\
AugmentGAN & authors published articles in 2004 \\
Recomb. \cite{Jia2016DataRF} & authors whose articles published date is 2004 \\
 \hline
 & \textbf{select restaurant where cuisine = thai and takeout = True}  \\
 \hline
Ground Truth & thai restaurants that have takeout \\
AugmentGAN & restaurant has thai cuisine and takeout \\
Recomb. \cite{Jia2016DataRF} & restaurant that cuisine is thai and has takeout \\
 \hline
 & \textbf{select meal where restaurant = (select restaurant where star = 3)}  \\
 \hline
Ground Truth & what is a meal served at a three star rated restaurant \\
AugmentGAN & meal served at 3 star restaurant \\
Recomb. \cite{Jia2016DataRF} & meal that restaurant whose star rating is 3 \\
 \hline
\end{tabular}
\end{adjustbox}
\end{center}
\caption{Examples of NL queries generated by AugmentGAN and recombination \cite{Jia2016DataRF}, compared with ground truth composed by crowdsourcing \cite{Wang2015BuildingAS}.}
\label{table:gan_examples}
\end{table*}

We first train a two-layer CTN for column tagging, and then value tagging is based on the pretrained two-layer column CTN with one extra layer on top; 
by doing so the value alignment during value tagging is improved, given that the pretrained two layers provide important information related to columns. 
For both encoder and decoder in seq2PSQL, we use 2-hidden-layer GRU cells \cite{chung2014empirical} and hidden states with size $256$; 
dropout for both encoder and decoder is applied during training with keep-rate $0.7$ for input and $0.5$ for output.
The decoder is based on beam-search with a beam size of $5$.

We conduct ablation analysis (CTN and seq2PSQL) to demonstrate the performance of SQIN, and compare with previous works on in-domain tasks.
From Table~\ref{table:ablation}, our model exhibits a better performance on all three datasets: 
seq2PSQL alone without a CTN demonstrates a better structural-aware decoder;
to demonstrate the performance of CTN, we evaluate both SQIN and a combined model CTN+seq2seq, which feeds the tagged input $P$ into a seq2seq model \cite{Jia2016DataRF};
CTN significantly enhances the performance of seq2seq by separating the schema-related information from the NL inputs, 
and with a better structural-aware decoder (seq2PSQL), SQIN shows a state-of-the-art performance for in-domain semantic parsing.

\subsection{Data Augmentation and Evaluation}

The generator is first pre-trained by predicting the NL queries given the SQLs based on maximum-likelihood-estimation (MLE) loss. 
The discriminator is also pre-trained: half of the negative examples we use are partial NL queries with incomplete information of corresponding SQLs; 
a quarter of the negative examples are complete NL queries with sequence being randomly permutated; the other quarter is generated from sampling. 

In Table.~\ref{table:gan_examples} we show several generated examples in \textsc{Overnight} by AugmentGAN and recombination method \cite{Jia2016DataRF}, with comparison to the ground truths, which are originally composed through crowdsourcing \cite{Wang2015BuildingAS}. From Table.~\ref{table:gan_examples}, the examples generated by recombination have more strict rules, whereas examples by AugmentGAN are more flexible in both sentence structure and words selection, thereby more resemble to human usage.

\begin{table}[ht]
\begin{center}
\begin{adjustbox}{center}
\begin{tabular}{ c | c | c | c }
 \hline
Subdomain & GAN & Tie & Ground truth \\
 \hline
\textsc{Restaurants} & 8 & 28 & 64 \\
\textsc{Publication} & 7 & 24 & 69 \\
 \hline
\end{tabular}
\end{adjustbox}
\end{center}
\caption{Crowd-sourced evaluation of $100$ randomly selected (GAN-generated, ground truth) NL pairs on each of two \textsc{Overnight} subdomains. The values represent the number of examples picked to be more human-like usage.}
\label{table:pairwise}
\end{table}

In order to qualitatively evaluate how good the NL queries are generated from AugmentGAN, we employ crowd-sourced judges to evaluate 100 randomly sampling pairs of human and GAN-generated queries. 
For each pair of queries, we ask 3 judges to decide which one is better, with tie allowed.
A small set of $20$ pairs is used to validate the quality of the crowd-sourced annotations, 
where only one annotator passes the validation set with a $95\%$ correctness could his annotations be accepted.

In Table.~\ref{table:pairwise} we show the crowd-sourced evaluation of the GAN-generated NL queries versus the ground truth in two \textsc{Overnight} subdomains; around $30-35\%$ of the generations resemble (or better than) human usage. For industrial purpose, even if the AugmentGAN cannot generate $100\%$ human-like queries, an extra selection step could be added; for most circumstances, the action of selecting is significantly less time-consuming than human directly composing or paraphrasing a NL query. Therefore, AugmentGAN exhibits both academic and practical impacts.

\begin{table*}[ht]
\small
\begin{center}
\begin{adjustbox}{center}
\begin{tabular}{l|c|c|c|c}
\hline
Domain-Transfer Approaches & seq2seq \cite{su2017cross} & seq2PSQL & CTN + seq2seq & SQIN \\ 
\hline
In-domain  & $.783$ & $.815$ & $.819$ & $.827$ \\
Plain transfer   & $.241$ & $.325$ & $.630$ & $.684$ \\
Limited target-domain data   & $.739$ & $.737$ & $.744$ & $.746$\\
Limited target-domain data + ReComb. \cite{Jia2016DataRF}   & $.751$ & $.763$ & $.786$ & $.795$ \\
Limited target-domain data + GAN   & $.793$ & $.800$ & $.829$ & $.835$ \\
Massive target-domain data   & $.832$ & $.838$ & $.844$ & $.854$ \\
\hline
\end{tabular}
\end{adjustbox}
\end{center}
\caption{Domain Adaptation Evaluation on \textsc{Overnight} (\textsc{Blocks} excluded).
We compare (vanilla seq2seq, seq2PSQL, CTN + seq2seq, SQIN) on five domain-transfer approaches.}
\label{table:in-domain-cross-domain-performance}
\end{table*}

\subsection{Domain Adaptation and Evaluation}
We evaluate how well our models (SQIN, seq2PSQL without tagging, and CTN + seq2seq) could leverage on the learning from source-domain data 
to generate SQL queries against target domain in dataset \textsc{Overnight}, compared with vanilla seq2seq (Table~\ref{table:in-domain-cross-domain-performance}). 

(1) \textit{in-domain} setting refers to the model both trained and tested on the target domain; 
(2) \textit{plain transfer} setting directly applies trained model on target-domain, where the source tables are all the \textsc{Overnight} subdomains except the target domain;
(3) \textit{massive target-domain data} uses sufficient amount of target-domain data to fine tune the model;
(4) \textit{limited target-domain data} uses randomly select $1/5$ of the target-domain data to fine tune the model; 
(5) \textit{limited target-domain data + GAN} refers to transfer with limited target-domain data with augmentation by GAN; 
(6) \textit{limited target-domain data + ReComb.} refers to transfer with limited target-domain data with augmentation by recombination \cite{Jia2016DataRF}.
For approaches with data augmentation, the size of augmented data is a factor of $4$ of seed data. 

 
From Table~\ref{table:in-domain-cross-domain-performance}, \textit{massive} performs better than \textit{in-domain}, 
which illustrate the out-of-domain information could enhance learning \cite{su2017cross}.
Both models with schema tagging (SQIN and CTN + seq2seq) have better results than their non-tagging counterparts, 
as well as demonstrate a better \textit{plain transfer} performance, 
showing that the separation of schema information selectively enhances learning of general NL knowledge and provides a better domain adaptability. 

For approaches using augmentation, AugmentGAN is more effective than recombination \cite{Jia2016DataRF}, by showing a better performance for all models. 
Using AugmentGAN, the accuracies for all models are higher than those of \textit{in-domain} setting, and even close to \textit{massive} setting, 
showing that even with limited target-domain data, good domain adaptation is still achieved.
One interesting thing to note: for the cases using AugmentGAN, 
even though there are two thirds less human-like queries in the training data, 
with the help of schema tagging, the models are still able to generate high-accuracy SQLs, 
which implies that a satisfactory transfer learning doesn't require completely human-resemble augmentation.

\begin{figure}[ht]
    \includegraphics[width=0.465\textwidth]{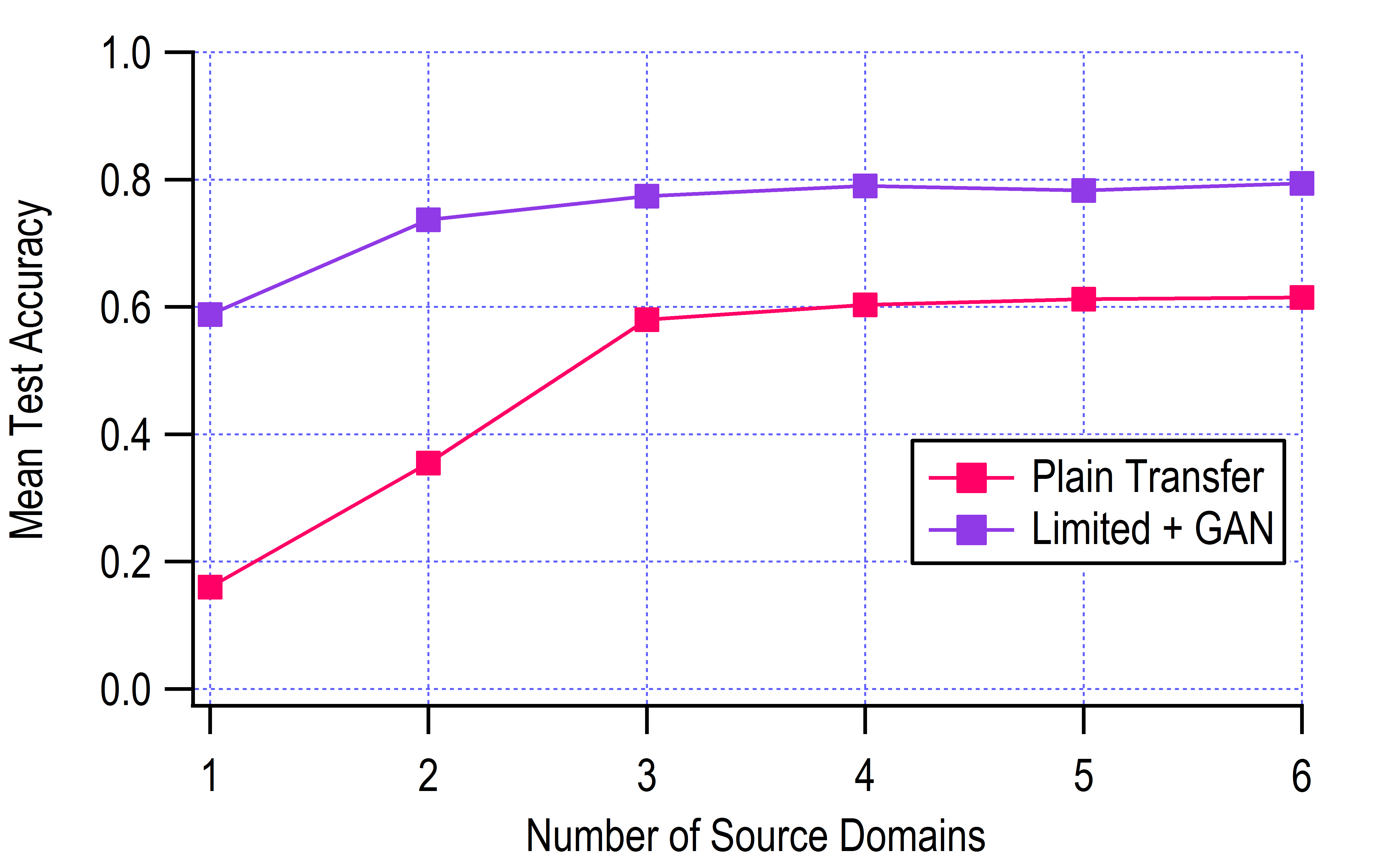}
    \caption{SQL accuracies on \textsc{Publication}, transferring from different number of source domains, with \textit{plain transfer} (pink) and \textit{limited.+ GAN} (purple) approaches.}
    \label{fig:5to1}
\end{figure} 

In the end, we evaluate how many source domains are needed for the model to generate correct SQL on target domain: 
we use different number of \textsc{Overnight} subdomains as the source tables, and subdomain \textsc{Publication} as the target table, 
and calculate the SQL generation accuracies for both \textit{plain transfer} and \textit{limited target domain data + GAN} approaches.

From Fig.~\ref{fig:5to1}, there are two observations: 
(1) if the source tables do not fully cover all possible queries types on target table, 
fine-tuning from target domain data is necessary to achieve a better saturation performance: i.e. self-join type of queries are included in subdomain \textsc{Publication} but not in other subdomains; 
(2) for a model previously trained on a sufficient number of source tables ($>3$ in this case), 
it is enough to feed a small amount of target domain data (+GAN) to achieve a good domain adaptation, 
a promising technique that could save resources and man-power when adapting to a new table.

\section{Conclusion \& Perspective}
\label{sec:conclusion}
As one of our main insights, we developed SQIN to separate schema-related information from the NL inputs, 
which enhances the learning of sequence-based models on general NL knowledge from source domains, 
thereby improving the in-domain performance and cross-domain adaptability.
Based on recombining a formal language (SQL) and correspondingly generating NL texts, 
we develop an effective GAN-based data augmentation algorithm, 
which could significantly reduce the human effort for composing data. 
Our extensive experiments demonstrate the advantage of our approaches.
Our extensive experimental analyses demonstrate the effectiveness of our approach on standard datasets. 
Future work could be 
to extend to other types of structured data by combing syntax-directed generation \cite{dai2018syntaxdir}.

\bibliographystyle{aaai}
\bibliography{IEEEfull}

\end{document}